\def\year{2020}\relax
\documentclass[letterpaper]{article} 
\usepackage{aaai20}  
\usepackage{times}  
\usepackage{helvet} 
\usepackage{courier}  
\usepackage[hyphens]{url}  
\usepackage{graphicx} 
\usepackage{graphicx}  
\usepackage{cuted}
\usepackage{amsmath}

\usepackage{subfig}

\title{Video Person Re-ID: Fantastic Techniques and Where to Find Them}

\author{\\  \Large \textbf{ Priyank Pathak,\textsuperscript{\rm 1}\thanks{Work done at Clarifai Inc.} Amir Erfan Eshratifar,\textsuperscript{\rm 2}\footnotemark[1]  Michael Gormish\textsuperscript{\rm 3}\footnotemark[1]} \\ 
\textsuperscript{\rm 1} Department of Computer Science, New York University,
New York, NY 10012, USA \\ 
\textsuperscript{\rm 2}University of Southern California, Los Angeles, CA 90089, USA\\
\textsuperscript{\rm 3}Clarifai, San Francisco, CA 94105, USA\\
ppriyank@nyu.edu, eshratif@usc.edu, michael.gormish@clarifai.com
}

\begin{document}

\maketitle

\begin{abstract}
The ability to identify the same person from multiple camera views without the explicit use of facial recognition is receiving commercial and academic interest. The current status-quo solutions are based on attention neural models. In this paper, we propose Attention and CL loss, which is a hybrid of center and Online Soft Mining (OSM) loss added to the attention loss on top of a temporal attention-based neural network. The proposed loss function applied with bag-of-tricks for training surpasses the state of the art on the common person Re-ID datasets, MARS and PRID 2011. Our source code is publicly available on github\footnote{https://github.com/ppriyank/Video-Person-Re-ID-Fantastic-Techniques-and-Where-to-Find-Them}. 
\end{abstract}

\section{Introduction}
\begin{figure}[!t]
 \centering
 \includegraphics[width=.95\columnwidth]{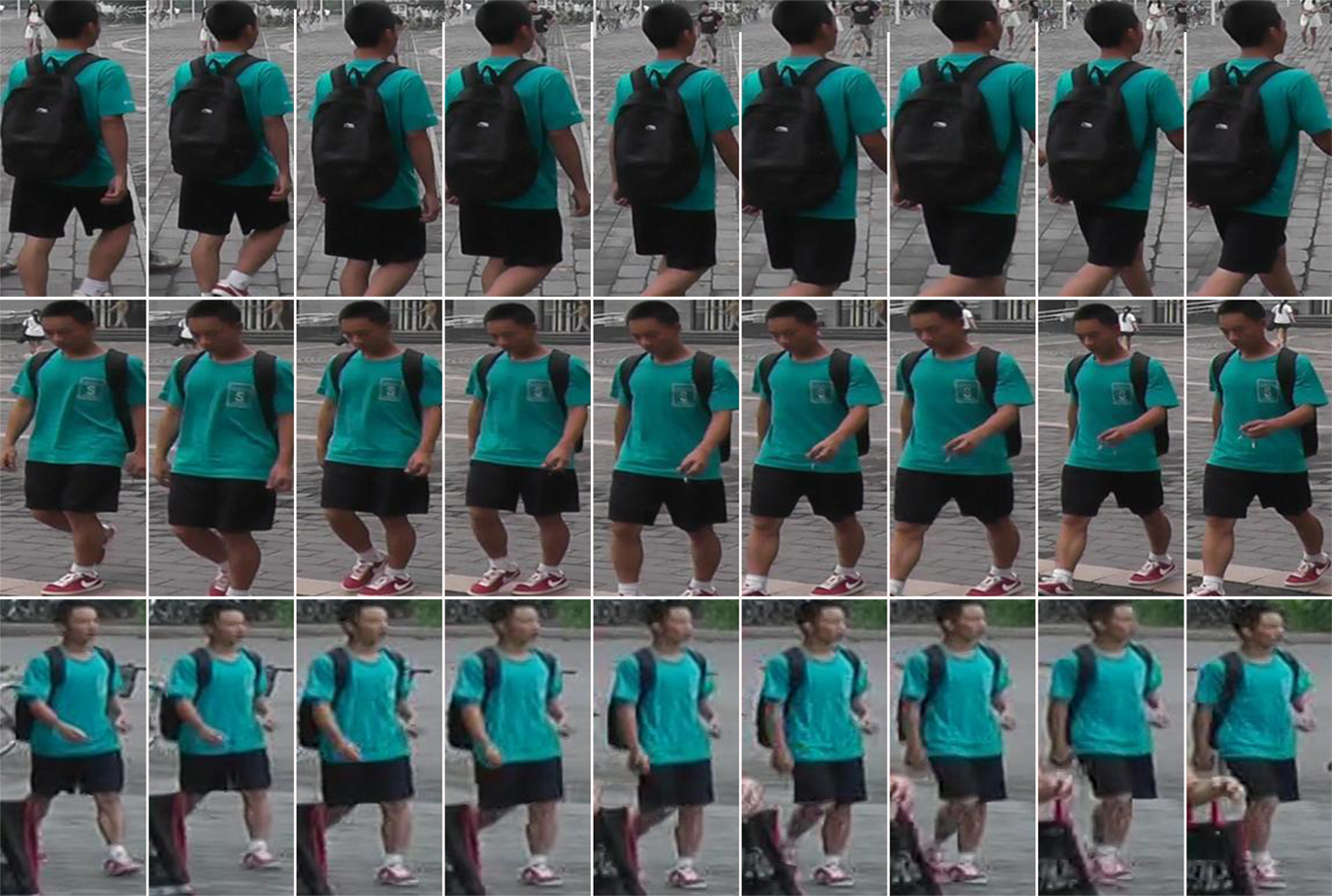}
 \caption{Data Samples are taken from  MARS \cite{MARS}. 
 Same individual captured in 3 input video clips
}
 \label{fig:data}
 \end{figure}

\noindent  Person Re-IDentification (Re-ID) aims to recognize the same individual in different pictures or image sequences caught by distinct cameras that may or may not be temporally aligned (Fig \ref{fig:data}). The goal of video-based person Re-ID is to find the same person in a set of gallery videos from a query video. One industrial use case is surveillance for security purposes. In contrast with image-based person Re-id, which can be consequently affected by several factors such as blurriness, lighting, pose, viewpoint and occlusion, video-based person Re-ID is more robust to these pitfalls as multiple frames distributed across the video are used.

\section{Methodology}

\begin{figure*}[!ht]
 \centering
 \includegraphics[width=0.99\textwidth, ]{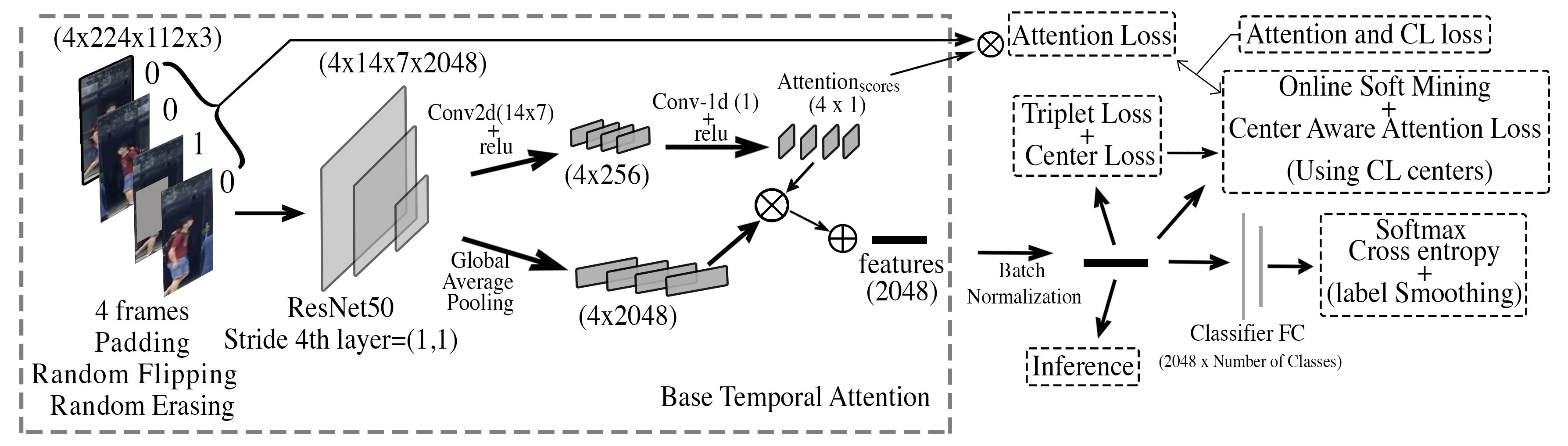}
 \caption{The proposed model architecture. $\otimes$ indicates pairwise multiplication and $\oplus$ indicate summation. 
}
 \label{fig:architecture}
 \end{figure*}
 
\textbf{Baseline (Base Temporal Attention): }
We use \citeauthor{baseline}'s \shortcite{baseline} temporal attention model as our base model, where a pre-trained ResNet-50 on ImageNet creates features for each  frame of a video clip, and an attention model computes a weighted sum of the features across frames.

\begin{table*}[!ht]
\centering
\begin{tabular}{|| p{6.4cm} || c | c | c | c || }

  \hline
  Model & mAP (re-rank)& CMC-1 (re-rank)& CMC-5 (re-rank) & CMC-20 (re-rank) \\\hline

  SOTA w/o re-ranking (\citeauthor{mars_sota}, \shortcite{mars_sota}) & 81.2 (-) & 86.2(-) &  95.7(-) & - (-) \\  \hline
  SOTA with re-ranking (\citeauthor{mars_sota}, \shortcite{mars_sota}) & 80.8 (87.7) & 86.3(87.2) &  95.7(\textbf{96.2}) & 98.1 \textbf{(98.6)} \\  \hline
  Baseline (\citeauthor{baseline}, \shortcite{baseline}) &  76.7 (84.5) & 83.3 (85.0) &  93.8 (94.7) & 97.4 (97.7) \\  \hline
  Baseline + Bag-of-Tricks (B-BOT)* & 81.3 (88.4) & 87.1 (87.6) & 95.9 (96.0) &  \textbf{98.2} (98.4) \\ \hline
  B-BOT + OSM loss (B-BOT + OSM)*
  & 82.4 (88.1) & 87.9 (87.6) & 96.0 (95.7) & 98.0 (98.5) \\\hline
\textbf{(Proposed)}  B-BOT + OSM +  CL Centers*
  & 81.2 (\textbf{88.5})  & 86.3 (\textbf{88.0}) & 95.6 (96.1)& \textbf{98.2} (98.5)  \\\hline
  \textbf{(Proposed)}  B-BOT + Attention and CL loss*
  & \textbf{82.9}(87.8)  & \textbf{88.6(88.0)} & \textbf{96.2} (95.4)& 98.0(98.3)  \\\hline
                         
\end{tabular}
\caption{MARS Dataset Performance. '-' indicates the results were not reported. '*' refers to hyperparameter optimized.}
\label{tab:mars1}
\end{table*} 

\begin{table*}[!ht]
\centering
\begin{tabular}{|| p{10cm} ||  c | c | c || }
  \hline
  Model & CMC-1 & CMC-5 & CMC-20  \\\hline
  SOTA (\citeauthor{prid_sota}, \shortcite{prid_sota}) &  96.1 & 99.5 &  -  \\  \hline
  \textbf{(Proposed)}  B-BOT +  Attn-CL loss*
   & 93.3 & 98.9 & 100.0 \\\hline
   \textbf{(Proposed)}  B-BOT + Attn-CL loss (pre-trained on MARS dataset)*
   & \textbf{96.6} & \textbf{100} & 100.0  \\\hline
\end{tabular}
\caption{PRID 2011 Dataset Performance. '-' indicates the results were not reported. '*' refers to hyperparameter optimized.}
\label{tab:prid2}
\end{table*} 

\noindent \textbf{Bag-of-Tricks:}
\citeauthor{bagoftricks} \shortcite{bagoftricks} proposed a series of tricks to enhance the performance of a ResNet model in image-based person Re-ID, which includes reducing the stride of the last layer (richer feature space), using warm-up learning rate, random erasing of patches within frames (simulating occlusion), label smoothing, center loss in addition to triplet loss, cosine-metric based triplet loss (angular distance proven better than L-2 distance) and lastly, batch normalization before the classification layer. Our experiments reveal that batch normalized feature provides a more robust model compared to non-normalized features.

\noindent \textbf{Attention and CL loss:}
\citeauthor{osm_caa} \shortcite{osm_caa} proposed Online Soft Mining and Class-Aware Attention (OSM loss) as an alternative to triplet loss for training Re-ID tasks, a modified contrastive loss with attention to remove noisy frames. We propose \textbf{CL Centers} OSM loss, which uses the center vectors from center loss as the class label vector representations, for cropping out noisy frames as they have greater variance compared to the originally proposed classifier weights. In addition, we penalize the model for giving high attention scores to frames where we have randomly deleted a patch. Such randomly erased frames are labeled as 1 otherwise 0.
\begin{equation}
\text{Attention loss} = \frac{1}{N} \sum^N_{i=1} label(i) * Attention_{score}(i)    
\end{equation}
where N is the number of total frames. \\ The Attention loss combined with OSM loss and CL Centers is denoted as \textbf{Attention and CL loss}.

\noindent \textbf{Hyperparameter optimization:} We also applied Facebook's hyperparameter optimization tool\footnote{https://github.com/facebook/Ax} to do hyperparameter search.

\noindent \textbf{Datasets:} We focus on the MARS and PRID datasets containing 1251 and 178 identities, respectively which are equally split among training and testing sets.

\section{Evaluation and Results}
In our experiments, we use four frames of the video selected randomly, $N=4$. Figure \ref{fig:architecture} shows the proposed model architecture. Table \ref{tab:mars1} and Table \ref{tab:prid2} show a comparison of our model to the state-of-the-art results on MARS and PRID 2011 datasets.

\section{Conclusion and Future Work}
In this paper, we mixed and improved existing techniques to surpass the state-of-the-art accuracy on MARS and PRID 2011 datasets. We plan to evaluate our work on other datasets and other similar tasks like facial re-identification in the future. 


\bibliographystyle{aaai}
\bibliography{references}

\end{document}